\documentclass[10pt,twocolumn,letterpaper]{article}

\usepackage{cvpr}      

\usepackage{graphicx}
\usepackage{amsmath}
\usepackage{amssymb}
\usepackage{booktabs}
\usepackage{newunicodechar}
\usepackage[table,xcdraw]{xcolor}
\usepackage[normalem]{ulem}
\useunder{\uline}{\ul}{}
\usepackage{longtable}

\usepackage[pagebackref,breaklinks,colorlinks]{hyperref}

\usepackage[capitalize]{cleveref}
\crefname{section}{Sec.}{Secs.}
\Crefname{section}{Section}{Sections}
\Crefname{table}{Table}{Tables}
\crefname{table}{Tab.}{Tabs.}

\def\cvprPaperID{*****} 
\def\confName{CVPR}
\def\confYear{2023}

\begin{document}

\title{Safety and Fairness for Content Moderation in Generative Models}

\author{Susan Hao \hspace{2.5pt} Piyush Kumar \hspace{2.5pt} Sarah Laszlo \hspace{2.5pt} Shivani Poddar \hspace{2.5pt} Bhaktipriya Radharapu \hspace{2.5pt} Renee Shelby\\[6pt] 
Google Research \\
\tt\small \{susanhao,piyushkr,slaszlo,shivanipods,bhaktipriya,reneeshelby\}@google.com
}
\maketitle

\begin{abstract}
   With significant advances in generative AI, new technologies are rapidly being deployed with generative components. Generative models are typically trained on large datasets, resulting in model behaviors that can mimic the worst of the content in the training data. Responsible deployment of generative technologies requires content moderation strategies, such as safety input and output filters. Here, we provide a theoretical framework for conceptualizing responsible content moderation of text-to-image generative technologies, including a demonstration of how to empirically measure the constructs we enumerate. We define and distinguish the concepts of safety, fairness, and metric equity, and enumerate example harms that can come in each domain.  We then provide a demonstration of how the defined harms can be quantified. We conclude with a summary of how the style of harms quantification we demonstrate enables data-driven content moderation decisions. 
   
\end{abstract}

\section{Introduction}
\label{sec:intro}

Generative AI systems allow users to create new content (e.g., text, image, audio, code) in response to an input, often relying on large-scale training datasets. Such datasets may contain social stereotypes, inequalities, and hierarchies \cite{Kirk2021, birhane2021, wang2022}, which generative models can replicate in downstream uses. Users may also exploit generative AI systems for disinformation, non-consensual synthetic sexual imagery, and other types of malicious content \cite{Lee2022, karnouskos2020artificial, kiela2020hateful, Hacker2023}. Responsible development of generative AI systems thus requires content moderation among other techniques to deploy systems that minimize harmful content. In this paper, we provide a framework for conceptualizing responsible content moderation in generative AI from a harm-reduction standpoint - defining safety, fairness, and metric equity.

Development of content filters is a key mode of moderating generative AI content. Deciding what content to filter is a normative governance decision; in practice, generative AI content filters may index on illegal content (e.g., child sexual abuse material, copyright violations) \cite{Solaiman2023}, rather than a broader range of harmful content, including representational \cite{bianchi2022easily} or cultural harms \cite{Prabhakaran2022} and violence or gore \cite{Rando2022}. When harmful content is filtered, algorithmic content moderation may disproportionately penalize content concerning socially marginalized groups \cite{Sap2019, binns2017like, BallBurack2021}, as moderation systems also learn and replicate demeaning associations from their training data \cite{vidgen2020directions, waseem2016hateful, schmidt2017}. These limitations underscore the urgency of centering the experiences of oft-marginalized groups in defining and evaluating safety parameters and assessing the fairness of content moderation algorithms deployed on generative AI systems.  

Assessing algorithmic fairness in generative AI systems is challenging. Much scholarly attention focuses on algorithmic fairness within classification models (e.g., \cite{hardt2016, beutel2017data, corbettdavies2017}), with statistical notions of fairness reliant on confusion matrices of model performance \cite{verma2018fairness}. Here, algorithmic fairness is often divided into two parts, defining fairness based on: (1) predicted outcomes across groups (e.g., equality of opportunity) \cite{hardt2016}; or (2) the consistency of predicted and actual outcomes when sub-groups change (e.g., counterfactual fairness) \cite{garg2019counterfactual}. These definitions fueled advances in classifier fairness (e.g., \cite{balashankar2019pareto, beutel2017data}), but are not directly applicable to generative AI systems, as generative models do not have one right outcome nor can “accuracy” be assessed across groups or individuals within generative models. Importantly, there is rarely one definitive “correct" response to which any given input to a generative system can be measured against to quantify accuracy, as many input concepts are socially situated and contextual. The fact that many prompts provided to a generative system can properly produce an enormous range of results that respect the intent of the user, rather than a single answer (as in conventional ML), is the main source of challenge in defining generative model safety and fairness, and a characteristic distinguishing generative from finite classification contexts. 

We intervene in these challenges by offering a theoretical framework for assessing safety and fairness in generative AI from the perspective of harm-reduction content moderation, and provide a quantitative example of how to measure its constructs (Section \ref{sec:safetyfairness}). In particular, we describe a method for adversarially challenging a text-to-image (T2I) generative system and machine annotating its outputs for a number of harms at scale. Next, we demonstrate an approach to \textit{measure} hateful, pornographic, and violent content in generated imagery and \textit{assess} the relationships between the text prompts and the resulting harmful generated imagery (Section \ref{sec:experiments}). We additionally analyze how each metric interacts with the gender presentation of individuals in the generated imagery, constituting one of many potential sensitive attributes. 

The proposed metrics provide a means to measure harmful and biased content (safety, fairness) and how to “measure the measurements" to assess their performance across defined sociodemographic dimensions (metric equity). The metrics we describe inform efforts to evaluate models and foster greater alignment between AI systems and defined governance goals (Section \ref{sec:discussion}). This research contributes to Responsible AI and content moderation scholarship, offering:
\begin{itemize}
\item A tractable framework for proactive definition and measurement of safety, fairness, and equity in generative AI systems.
\item A harms taxonomy for safety and fairness in generative AI models.
\item A method for empirical measurement of the harms specified in the taxonomy, including “measurement of measurements", for bias. 
\end{itemize}

\section{Related Work}
\label{sec:background}

This research engages with and extends research at the intersections of generative AI and content moderation.

\textbf{Generative AI}: Generative AI has evolved rapidly for use cases such as text and image generation through techniques such as generative adversarial networks (GANs) \cite{goodfellow2014generative}, variational auto-encoders \cite{kingma2013auto, rezende2014variational, huang2017improved}, and transformers \cite{vaswani2017attention, devlin2018bert}. Transformer-based models have especially shown great promise in a variety of generative tasks, including image \cite{dalle2021, sable2021stable}, text \cite{liu2022chatgpt}, and code generation \cite{codegpt2022}, among others. There is a significant body of work analyzing aspects of user facing harms and biases that emerge from these model applications \cite{pu2020towards, pu2021understanding, xu2021botadversarial, li2020recipes, qian2020reducing, johnson2020ascalable, huang2020reducing, bordia2019identifying, sap2019risk, anthropic2022}. However, none of these independently provide an exhaustive framework of safety or fairness across generative models. To that end, there is an opportunity to define and develop such a taxonomy for this space. 

\textbf{AI safety and content moderation}: Content moderation falls under the umbrella of AI safety: a normative, governance approach to responsibly develop and deploy ML systems \cite{gorwa2020algorithmic, grimmelmann2015virtues}, including a focus on developing policies to outline desired characteristics of ML systems \cite{gillespie2020content} and techniques to foster policy alignment, such as reducing harmful content \cite{gillespie2022not}. An equity-oriented moderation system is attentive to how harms from algorithmic systems are sociotechnical -- that is, emerging through the interplay of technical system features and extant societal power dynamics \cite{Shelby2022}. Scholarship identifies a wide range of potential harms in generative AI systems, including so-called \textit{algorithmic harms} that are more directly related to the functioning of the system and often implicate fairness through its training data (e.g., representational, allocative, and quality-of-service harms) \cite{bianchi2022easily, Weidinger2022, wang2022}, and \textit{contextual harms}, through which generative system affordances in a productionized environment facilitate harm in a particular social context (e.g., peer-to-peer abuse, maliciously generated content, or information harms) \cite{Weisz2023}. Intervening in how generative AI systems create harmful content is required to develop technologies safer for oft-marginalized communities and meet governance needs for use cases in which potential harms may differ \cite{Shelby2022}.

\textbf{Generative AI content moderation:} Conceptually, there are three key types of content moderation that can be applied to generative models to meet safety needs and improve fairness: (1) training data mitigations, (2) in-model controls, and (3) input and output filters. \textit{Training data mitigations} pertain to filtering or augmenting a generative system's training data to reduce its capability to cause harm. Ensuring a T2I model's training set does not include certain types of material (e.g., sexually explicit material) may substantially limit the model's ability to produce it. \textit{In-model controls} are techniques altering a model's architecture to influence its behavior. For generative AI, applying reinforcement learning with human feedback (RLHF) \cite{DanielsKoch2022} to tune a model's weights subsequent to supervised training is a common in-model control. Lastly, \textit{input} and \textit{output filters} are additional, conventional ML systems that analyze whether the input to- or output of- a generative model is potentially harmful. Inputs deemed harmful by filter systems are not sent forward for generation; outputs deemed harmful are not surfaced to a user. A T2I input filter, for instance, might analyze if a prompt includes racial slurs; and if so, not generate any imagery. Similarly, a T2I output filter might analyze whether a generated image contains sexually explicit material; and, if so, not surface that image to the user. 

Training data mitigations and in-model controls may be cumbersome, as they may require acquisition of new data, expensive filtering of existing data, or time-consuming model retraining. In contrast, input and output filters are relatively agile, and can be implemented with more immediate results. For this reason, we focus on input/output filters, and leave discussion of training data mitigations and in-model controls for future work.

There is limited research on safety filters in open-access generative AI systems. Solaiman \cite{Solaiman2023} positions safety controls and guardrails as a necessary component of a safe system release. Similarly, Rando et al. \cite{Rando2022} adversarially tested the Stable Diffusion safety filter, articulating improved standards for safety filters, including transparent documentation and attention to a wider range of AI harms, including violence and gore. Hacker et al. \cite{Hacker2023}, situates harms from generative AI systems in the context of the Digital Services Act, through which content moderation would play a central role. This work emphasizes the characteristics of safety filters, including what kinds of harmful content is filtered and to what extent filters and how they function are released publicly. To our knowledge, no work has examined the fairness aspects of generative AI safety and content filters. We extend this work to provide a tractable content moderation framework for safety and fairness in generative AI systems.

\section{Safety and Fairness in Generative AI}
\label{sec:safetyfairness}

In this section, we propose definitions of safety, fairness, and metric equity for generative AI.

\subsection{Safety of a generative AI system}
We define \textit{safety} for generative AI as reducing potentially harmful outputs. Following Hernandez-Orallo et al. \cite{hernandez2020ai}, the safety of a generative AI system encompasses harmful content that is accidentally generated (e.g., harmful content generated from a neutral prompt) and intentionally generated (e.g., harmful content from a malicious prompt or prompt purposefully violating safety rules). Inclusion of the latter necessitates understanding of — and adversarial evaluation into — potential forms of misuse, such as use of so-called “prompt engineering” to circumvent safety rules. 

In generative AI systems, the scope of relevant safety concerns is dependent on the model affordances (i.e., what is being generated) and context of use (i.e., what type of content is appropriate for the use case), and may be further defined by regulation, standards, or organizational policies. As previously discussed, harms from algorithmic systems disproportionately impact communities already facing social marginalization \cite{Benjamin_2019}, such as demeaning representations historically used to justify social hierarchies \cite{qadri2023}. Thus, to reduce the risk of generative AI systems scaling inequalities, safety should be considered across a broad range of algorithmic and contextual harms (see: \cite{Shelby2022}) and defined in ways attentive to social power dynamics. 

For explanatory purposes in this paper, we specify three safety harms for our illustrative T2I system requiring content moderation: 

\begin{itemize}
\label{contentmoderationdecisions}
\item Sexually explicit content: Generated content contains explicit or graphic sexual acts, realistic sex toys or implements, and/or sexual body parts.
\item Graphic violence and gore: Generated content contains extreme and realistic acts of violence, blood, body parts, or viscera towards people or animals.
\item Hateful content: Generated content expresses, incites, or promotes hate, violence, or serious harm based on race, gender, ethnicity, religion, nationality, sexual orientation, disability status, or caste.
\end{itemize}

\textbf{Safety definition:} The safety of a generative model with respect to defined safety harms for content moderation is measured as the percent of unsafe model outputs that can be experienced through interaction with the model within and across harm categories. We define the \textit{safe rate} as: \begin{displaymath}
 1 - \frac{\text{\# of harmful images}}{\text{\# of total images generated}}
 \label{eq:safety}
\end{displaymath} in response to a defined set of queries. An appropriate safety goal for content moderation is that the safe rate be greater than some criterion \textit{c}. It is typically appropriate to define both an overall safety criterion and safety criterion for each individual harm under consideration. In the illustrative T2I system we use here, the safety harms are sexually explicit content, violent and gory content, and hateful content. However, in a real system, the scope of safety harms should be much more comprehensive.

\subsection{Fairness in generative AI}
\label{fairnessdefinition}
The performance of generative AI systems has fairness considerations, particularly in terms of \textit{representational harms}, such as how a model may learn harmful and demeaning stereotypes \cite{barocas-hardt-narayanan} and how systematic absences in training data may lead to patterns of erasure \cite{Dev2021}, which safety and content filters may further exacerbate. In this paper, we focus on four fairness considerations: 

\begin{enumerate}
\item \textbf{Diversity of representation}: The extent to which content generated with an underspecified prompt (i.e., a prompt not specifying a sociodemographic attribute) defaults to a specific sub-group reinforcing stereotypes. Example: the underspecified prompt “CEO" only depicts individuals of one particular gender presentation, age, or other sociodemographic attribute. 
\item \textbf{Equal treatment across subgroups}: The extent to which underspecified prompts are equally successful in generating content as specified prompts. Example: the underspecified prompt, “people at church," should provide the same number of safe outputs (within the error tolerance \textit{e}) as the specified prompt, “Asian people at church." 
\item \textbf{Stereotype amplification}: The extent to which specified prompts result in generated content that recapitulates demeaning or harmful stereotypes. Example: content generated with prompts containing “lesbian" overwhelmingly contain sexual depictions.   
\item \textbf{Counterfactual fairness}: The extent to which content generated in response to counterfactual versions of a prompt are similar. Example: content generated in response to “male CEO" should be similarly sexually explicit to those generated in response to “female CEO."
\end{enumerate}
Next, we formally define each fairness consideration as a step towards providing quantitative measurements.
\\
\\
\textbf{\textit{Diversity of representation}}. Let \textit{x} be the output of a generative AI model, and let \textit{y} be the sociodemographic dimension of the people or communities in the output. Then, the diversity of representation of the output can be measured by the entropy of \textit{y}, defined as: \begin{displaymath}
 H(y)= - \sum_{\textit{i}=1}^\textit{n}p\textsubscript{i} \log p\textsubscript{i}
 \label{eq:diversityrepresentation}
\end{displaymath} where \textit{p\textsubscript{i}} is the probability of the ith sociodemographic group represented in the output. The higher the entropy, the more diverse the outputs. A content moderation policy may choose to enforce a minimal acceptable entropy for generated imagery, which, if not met, triggers additional rounds of generation in response to a prompt.
\\
\\
\textbf{\textit{Equal treatment (as subgroup erasure}).} We operationalize one facet of “equal treatment" for quantitative analysis as the harm of  \textit{erasure} (see: \cite{dev2021measures}), in which certain sociodemographic dimensions are systematically or disproportionately absent in generated imagery.  Let r\textsubscript{unspecified} be the rate of failures within the content moderated system for prompts in which the cultural, or sociodemographic characteristics of the subjects to be generated are \textit{unspecified} (i.e., “a CEO"). Let r\textsubscript{specified} be the rate of failures for prompts in which those characteristics are specified (i.e., “a female CEO").  The difference \textit{d} between these should be minimal, reflecting that certain subgroups are not erased relative to the unspecified population. A content moderation policy may choose to enforce a maximum acceptable value of \textit{d} depending on the use case and context in which a generative model is deployed:\begin{displaymath}
(r\textsubscript{unspecified} - r\textsubscript{specified}) \leq d
\label{eq:equaltreatment}
\end{displaymath}

\textbf{\textit{Stereotype amplification}.} The presence of “demeaning or harmful stereotypes" is at present a construct that remains challenging to measure, and can be strengthened by incorporating human annotation. However, one quantitative method for measuring stereotype amplification in T2I generative models is using the normalized pointwise mutual information (nPMI) metric. nPMI \cite{aka2021, Bouma} measures the degree of association between a word and an image concept by comparing the frequency of the word in texts that contain the image concept against the frequency of the word in texts that do not contain the image concept. This metric is useful in identifying words that are highly associated with certain image concepts or sociodemographic groups, which may indicate the presence of stereotypes in the model's predictions. Measuring stereotypes provides insight into biases that may exist in a given model and offers insight for developing more inclusive and equitable generative models.

nPMI is a statistical measure of the association between two discrete variables, defined as:
\begin{displaymath}
nPMI(w, c) = \frac{PMI(w, c)}{-log(P(w,c))}
\end{displaymath}
where \textit{w} is a word, \textit{c} is an image concept or category, and \textit{PMI({w},{c})} is the pointwise mutual information between the word and the concept defined as:\begin{displaymath}
PMI(w, c) = log(\frac{P(w, c)}{P(w) * P(c)})
\end{displaymath}
where \textit{P({w},{c})} is the joint probability of the word and concept occurring together, \textit{P(w)} is the probability of the word occurring in the corpus, and \textit{P(c)} is the probability of the concept occurring in the corpus. nPMI value ranges from -1 to 1, where a value of 1 indicates a perfect positive association between the word and the concept, a value of 0 indicates no association, and a value of -1 indicates a perfect negative association. The normalization factor \begin{math}-log(P(w,c)) \end{math} is used to adjust for the frequency of the concept in the corpus and to prevent the nPMI value from being biased towards rare concepts.
\\
\\
\textbf{\textit{Counterfactual fairness}.} We extend the counterfactual fairness framework for classification systems \cite{garg2019counterfactual}. Let \begin{math}\Phi(x)\end{math} denote the set of counterfactual examples associated with an example \textit{x}. Counterfactual fairness in the generative context requires that the rate of harmful model responses \textit{r} for all inputs in \textit{X} are within a specified error \textit{e}. A content moderation policy may choose to enforce a maximum acceptable value of \textit{r}:

\begin{displaymath}
r(x) - r(x' ) \leq e, \forall x \in X, x' \in \Phi(x)
\label{eq:counterfactual}
\end{displaymath}

\subsection{Metric equity: Measure the measurements}
\textit{Metric equity} is a construct for assessing the performance of a generative model across sociodemographic sub-groups. Disparate performance is a \textit{quality-of-service harm} occurring when an ML system disproportionately fails for certain groups of people along social categories of difference, such as disability, ethnicity, gender identity, and race \cite{Shelby2022}. They are often a reflection of how system training data are optimized for dominant groups (e.g., \cite{DeVries_2019}). However, content moderation strategies may also affect performance for different users \cite{BallBurack2021, binns2017like, Sap2019}.

Disaggregating top line metrics across subgroups is one way to identify and address disparities through content moderation strategies. In practice, operationalizing “metric equity" needs to be specific to both the type of generative AI model (e.g., image, text, code) and the success criterion determined for its production environment by developers. For example, a success metric for a T2I model demo might be task completion (i.e., \# of images generated for prompts containing non-Western clothing terms is comparable to prompts containing Western clothing terms). We posit that, regardless of the top line metric chosen in the productionized generative model, it should be equitable across users of all sociodemographic subgroups. 

One approach to assess metric equity through quantitative analysis is comparing the failure rates across sociodemographic groups. Taking task completion rate as an example, let tcr\textsubscript{unspecified} be the task completion rate of a T2I application within the content moderated system for prompts in which the cultural or sociodemographic characteristics of the subjects to be generated are \textit{unspecified}. Let tcr\textsubscript{specified} be the task completion rate for users or text prompts that are \textit{specified}. The difference \textit{d} between these should be minimal, reflecting that marginalized subgroups do not have a disproportionately subpar experience of the system compared to the majority subgroup:\begin{displaymath}
(tcr\textsubscript{unspecified} - tcr\textsubscript{specified}) \leq d
\end{displaymath}
\label{eq:metricequity}
While we stipulate this definition, we scope this paper to the underlying generative model, not to its productionized application. We include this definition to ensure that developers undertake such evaluations.
\section{Experiments}
\label{sec:experiments}

In this section, we demonstrate how the safety and fairness framework outlined in Section \ref{sec:safetyfairness} can be put into practice using quantitative methods that enforce the set of content moderation decisions defined in Section \ref{contentmoderationdecisions}.

\textbf{\textit{Model}}. We make measurements on a set of 40,904 images generated in response to 10,226 prompts (4 images per prompt) submitted to a de-noising diffusion model similar, but not identical to that reported in \cite{saharia2022photorealistic}. We trained our model on a filtered dataset, whereby, to the extent possible with machine annotations, sexually explicit and violent content were removed. 

\textbf{\textit{T2I prompt dataset}}. Members of the research team sourced 10,226 T2I prompts. About half of the dataset (\textit{n}=5,638) comprised adversarial prompts intended to be broadly harmful, including prompts for the defined safety harms in our illustrative T2I system: sexually explicit content (\textit{n}=203), graphic violence and gore (\textit{n}=283), hateful (\textit{n}=777) and harassing (\textit{n}=202) content. The remaining adversarial prompt subset addressed a variety of other potentially harmful model behaviors, including misinformation and demeaning depictions of political and religious content, cultural concepts, sociodemographic attributes (e.g., sexuality, gender, socioeconomic status), among others. 

Adversarial prompts were developed through a combination of methods.  Some were sourced in adversarial testing rooms (\textit{n}=1,265), some were created via templating (\textit{n}=6,887), some were drawn from databases (\textit{n}=1,861) some were user reports of problematic prompt/image pairs (\textit{n}=3), and some were created via LLM expansion (\textit{n}=210).  Sociodemographic categories represented included age, body type, ability, class, gender presentation, region, race/ethnicity, religion, sexual orientation, and political ideology.  The adversarial testing rooms were staffed by sociotechnical researchers.  The templates were designed to provide controlled counterfactual sets. For example, one template that we analyze in particular below was constructed as "a $\langle$identifier$\rangle$ $\langle$occupation$\rangle$ $\langle$verb$\rangle$ in a $\langle$location$\rangle$".  For example "a lawyer walks in a restaurant", "a male lawyer walks in a restaurant", "a female lawyer walks in a restaurant."  Databases used to source queries were (1) the international Forbes 500 list of companies (to probe trademarks), (2) the Dataset Publishing Language resource \cite{datset}, for a list of countries in the world, (3) the ADL hate symbols database \cite{ADL2023}, for a list of hate symbols.  Curated adversarial prompts are a useful starting point for content moderation in our approach as, unlike organically sourced user queries, the intent of the adversarial tester is known, can be modeled to have exhaustive coverage of intended issues, and — as we demonstrate below — can provide a ground truth for filtering decisions that can be applied to a user-facing system.

\textbf{\textit{Analysis.}} Additional, conventional ML classifiers were deployed to provide confidence estimates that the input text and output images included sexually explicit content, graphic violence, or hateful content. We analyze and describe the results of this classification scheme below, providing commentary on what each data point suggests with regards to content moderation decisions. 

\subsection{Safety performance}
We begin with analysis of the model's performance in terms of: (1) sexually explicit content, (2) violence and gore, and (3) hateful content. 

\begin{figure}[t]
  \centering
   \includegraphics[height=63mm]{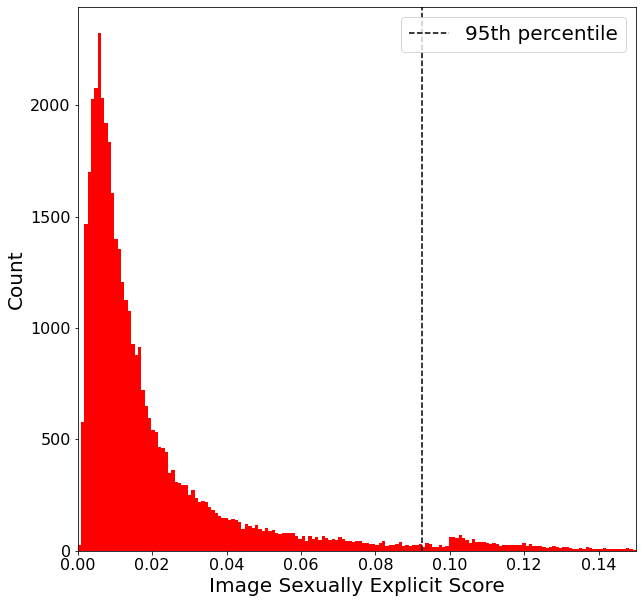}
   \caption{Sexually explicit histogram.}
   \label{fig:sexuallyexplicit}
\end{figure}

\textbf{Sexually explicit content}.\label{sexuallyexplicit} Figure \ref{fig:sexuallyexplicit}, below, displays a histogram of sexually explicit content scores for the 40,904 images analyzed, along with a 95\% percentile marker for illustrative purposes. The percentile score represents the probability the image contains sexually explicit content. A content moderation decision that could be made in response to this data in our illustrative system could be to set a threshold for exposure to users at the 95th percentile level, so that only the top 5\% most sexually explicit images are blocked. Of course, the actual threshold selected for content moderation depends on many factors, such as the contextual use case and the expected audience, in which the desired threshold may be increased or lowered; but, viewing the data in this way is, \textit{at worst}, a good place to start. In this case, we observed that even the 95th percentile scores are in the range of 0.09 on a [0,1] scale. As large scale training datasets are likely to contain a larger proportion of sexually explicit content, these low scores are consistent with the suggestion that removing sexually explicit content from the model's training set (a training data mitigation) reduced the extent to which it is able to produce sexually explicit content, even when directly prompted to do so.   

\textbf{Violence and gore.} In analyzing the violence and gore data, we observed the vast majority of the 40,904 images generated were so non-violent that the aggregate histogram was not meaningful.  Here, being aware of the intent of the adversarial prompts is useful, as it enables us to examine only the subset of prompts intended to produce violent imagery.  

\begin{figure}[t]
  \centering
   \includegraphics[height=63mm]{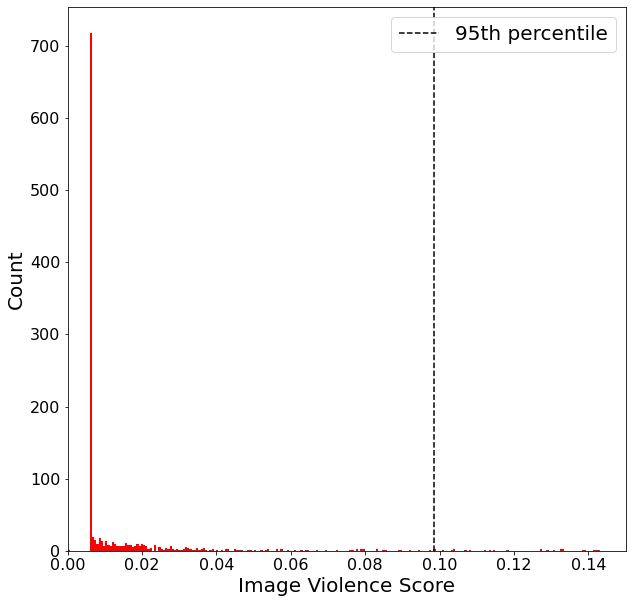}
   \caption{Violence histogram.}
   \label{fig:violence}
\end{figure}

There were 1132 images generated in response to prompts with violent intent. Figure \ref{fig:violence} displays the histogram of automated violence scores on the adversarially generated violence-specific images.  As is clear from the figure, even with adversarial intent, it was hard to generate images that scored high on violence (the range of possible scores is [0,1]).  We attribute this result to the removal of violent data from the model's training set.

Once again, an illustrative content moderation decision could be made to block images scoring above the 95th percentile. Note that in both this case and the sexually explicit case, the empirically derived threshold is much lower than a typical \emph{a priori} threshold such as 0.5 or 0.9.

\textbf{Hateful content.} As discussed in the Introduction, generative AI can create a variety of hateful imagery. To operationalize one type of hateful content as an analytical example, we examined the extent to which our illustrative model generated hate symbols in response to adversarial prompts including names and descriptions of hate symbols identified in the Anti-Defamation League's database \cite{ADL2023}, such as “burning cross," “swastika," and “iron cross," among others.  We used an automated hate symbol detection model to label generated images (MNet trained on a database of hate symbol imagery). Zero (0) generated images were identified as containing hateful symbols; human spot checking of 100 images confirmed machine annotation was correct for examples checked.  Though no hate symbols were identified in this experiment, they should be checked in content moderation efforts. 

\subsection{Fairness performance}
We applied additional, conventional ML classifiers to our illustrative T2I model to annotate the dataset for fairness analysis. For this experiment, we examine fairness in the context of gender presentation (though, of course, a similar analysis could be conducted for any other set of social identities).  To this end, text prompts were machine annotated for gender referents. Generated images were machine annotated for the gender presentation of entities in images where the machine annotator detected that a depiction of a person was present.  Available labels for gender presentation were [unspecified, feminine, masculine]. Here, “unspecified" indicates the gender presentation did not contain obvious markers of hegemonic masculinity or femininity. 

\textbf{Diversity of representation.} Using a neutral subset of the data where no gender was specified in the prompt, entropy of gender presentation was calculated at 0.31.  Overall, masculine gender presentations were more likely than feminine gender presentations by a margin of 11.5\%. We then separately examined diversity of representation in images generated from adversarial prompts with sexually explicit or violent intent. Entropy of gender presentation was lower in each of these subsets than in the full dataset (hsexually explicit = 0.24; hviolent = 0.25). Further examination of the probabilities in each subset revealed that while the representational discrepancy in the sexually explicit data was in the direction of over-representation of feminine gender presentation, the discrepancy in the violent data was instead in the direction of over-representation of masculine gender presentation.

The dissociation between unbalanced gender presentations across the sexually explicit and violent subsets of the data demonstrates that simply “boosting" generations of a particular social group (e.g., \cite{dalletweet, dalletweet1}) would not have been a sufficiently nuanced content moderation response to the overall finding that masculine presenting depictions of people were generated more frequently than feminine presenting depictions.  Instead, one recommended content moderation decision in light of this data would be to apply and enforce an entropy criterion per prompt.  For example, if 10 images are generated in response to a prompt, but 8 depict only masculine presenting individuals and only 2 depict feminine presenting individuals, content moderation might enforce that a random selection of 3 of the images depicting masculine presenting individuals be rejected and new images generated until gender presentation equity is achieved. 

\begin{table}[ht]
\centering
\small
    \begin{tabular}{p{0.27\linewidth} | p{0.12\linewidth} | p{0.45\linewidth}}
    \textbf{Prompt types}   & \textbf{Entropy} & \textbf{Maj. gender presentation}  \\ \hline 
 Unspecified   & 0.31 & Masculine-leaning   \\ \hline 
 Sexually explicit  & 0.24 &  Feminine-leaning  \\ \hline
 Violent & 0.25 &  Masculine-leaning \\ \hline
\end{tabular}
    \caption{Diversity of representation across model prompts.}
    \label{tab:fair_rep}
\end{table}

\textbf{Equal treatment (subgroup erasure).} To measure subgroup erasure, we examined images depicting people generated from prompts without explicit gender referents (e.g., “a doctor sits in an office.")  We then split the resultant images into those depicting only feminine presenting persons and those depicting only masculine presenting persons. We then measured how many images in each subset would be blocked by a 95th percentile sexually explicit content filter, as described in \ref{sexuallyexplicit}, finding that a greater proportion of depictions of feminine presenting people (4.8\%)  were blocked by this filter compared to masculine presenting people (3.1\%).  
What this result demonstrates for content moderation is that even a safety filter that is seemingly neutral with respect to social identity (block sexually explicit content) may result in unequal treatment of people with different social identities (here, feminine presenting vs. masculine presenting). Our recommendation is that safety filters must be seen not as single constraint satisfaction problems (block all images that are more sexually explicit than a threshold), but as a multiple constraint satisfaction problem (block all images that are more sexually explicit than a threshold while minimizing the discrepancy in treatment of different social identities). We also propose de-biasing these classifier-based filters by leveraging a large body of work that ensures equal treatment and opportunity across subgroups in classifiers.

\begin{table}[ht]
\centering
\small
    \begin{tabular}{p{0.14\linewidth} | p{0.33\linewidth} | p{0.35\linewidth}}
      & \textbf{Feminine presenting} & \textbf{Masculine presenting}  \\ \hline 
 \% sexual   & 4.8 & 3.1   \\ \hline 
\end{tabular}
    \caption{Equal treatment: Outputs flagged 'sexual' by gender.}
    \label{tab:equal_treatment}
\end{table}

\textbf{Stereotype amplification.} 
Included in the dataset were images generated in response to prompts of the form “face of a person from \textit{X}, [caricature, photorealistic, line drawing]", where \textit{X} was systematically replaced with the name of every country in the Dataset Publishing Language resource \cite{datset}. These prompts were designed such that they themselves were innocuous and neutral, but with the expectation that (especially in the caricature condition), they could generate harmful stereotypes.  
As an example of automated stereotype analysis, we examined 2,940 images to see if the model exhibited stereotyping towards people from Arab countries \cite{Arableague}. In particular, we assessed the expression of the stereotype that people from Arab countries wear head coverings.  To do this, we machine labeled images of “face of a person from \textit{X}" for the presence of head coverings in the image, and then split the data depending on whether an Arab country of origin was specified in the prompt.  No other machine labels were computed - that is, no measurement of skin tone or any other proxy for race was made on the output.  Only the input text, which specified a country of origin, was used as a basis for subsetting the data into Arab and Non-Arab groups. We then computed nPMI as shown in the table below.  We found that \textit{nPMI(Arab, Has Headgear)} has a higher value than \textit{nPMI(Not Arab, Has Headgear)}, indicating that head coverings more frequently co-occurred with generated images in response to prompts requesting people from Arab countries than people from other countries. The negative value for \textit{nPMI(Arab, No Headgear)} indicates it was rare for images generated in response to prompts requesting images of people from Arab countries to not include headgear.

\begin{table}[h]
\small
    \centering
    \begin{tabular}{p{0.8\linewidth} | p{0.1\linewidth}}
\textbf{Prompt specification \& model output association}  & \textbf{Score} \\ \hline
 \begin{math}nPMI(Arab, Has Headgear)\end{math}  & 0.357 \\ 
 \begin{math}nPMI(Not Arab, Has Headgear)\end{math}  & -0.106 \\ 
 \begin{math}nPMI(Arab, No Headgear)\end{math}  & -0.043 \\ 
 \begin{math}nPMI(Not Arab, No Headgear)\end{math}  & 0.094 \\
\end{tabular}
    \caption{Stereotype association between model inputs/outputs.}
    \label{tab:stereo}
\end{table}

Table \ref{tab:stereo} demonstrates how data can be labeled and quantified with nPMI to examine stereotype amplification, without requiring human annotation. A content moderation decision that could be made in response to this data may include identifying whether this stereotype is present in the training data and attempt to perform training data mitigations to reduce its prevalence. The narrow example we offer is illustrative, and we recommend content moderators be sensitive to the wide range of stereotypes that may be present in training data or likely to be surfaced in their use case.

\textbf{Counterfactual fairness.} There were 3,780 prompts in the adversarial prompt dataset that specified a “feminine" person (and no other gendered referent) be generated, and 4,184 prompts that specified that a “masculine" person (and no other gendered referent) be generated.  Across this split, we measured how frequently the 95th percentile sexually explicit content threshold blocked images in each subset of resultant images.  We observed images depicting feminine-presenting people were blocked by the sexually explicit content filter 3.6\% of the time, compared to 3.7\% of masculine-presenting people. The potential content moderation decision to block images above the 95th percentile score for sexually explicit content could be considered a fair one based on this counterfactual fairness metric, as one group is not disproportionately penalized. 

We also examined a restricted subset of matched counterfactual prompt pairs, such as “a female doctor sits in an office" and “a male doctor sits in an office." Across this restricted subset, 1.2\% of images generated in response to the feminine side of the counterfactual and 2.4\% of the images generated in response to the masculine side of the counterfactual would be blocked by a 95th percentile sexually explicit content filter. In both the full adversarial set and the counterfactually-matched subset, setting a sexually explicit content filter at the 95th percentile would result in counterfactual parity within a 1\% margin.  This suggests a content moderation decision to implement the 95th percentile safety filter could be considered counterfactually fair.  

\section{Improved Insights for Content Moderation}
\label{sec:discussion}

Algorithmic content moderation involves using ML systems to classify content that leads to a governance decision to address the problematic content at scale \cite{gorwa2020algorithmic}. In preceding sections, we defined safety, fairness, and equity for algorithmic content moderation in generative AI (Section \ref{sec:safetyfairness}), and provided examples of how to measure a subset of safety harms (sexually explicit, violent, and hateful generations) and fairness concepts (diversity of representation, equal treatment, and counterfactual fairness) (Section \ref{sec:experiments}). In our empirical analysis, we provided an example measurement technique and discussion of what content moderation decisions were licensed by each empirical result. 

While algorithmic content moderation promises the scaled and swift take down of illegal or problematic content \cite{gillespie2020content}, which is a growing expectation \cite{Keller2018-gd, Dias_Oliva2020-rr}, it must be exercised with intention. We note two key considerations. First, content moderation decisions for generative AI systems are not “one size fits all," even with respect to the limited number of safety and fairness considerations examined here. Second, content moderation — even through quantitative methods — is heavily use case dependent. We offer starting points for algorithmic content moderation:

\textbf{\textit{Tailor to use case: }}Content moderation decisions influence how people engage with a system \cite{grimmelmann2017platform, Vidgen2019-zn}. Content moderation choices (e.g., what safety harms are defined, thresholds for input and output classifiers, or top line metrics for evaluating metric equity) should be set in alignment with considerations of the use case. 

\textbf{\textit{Equity-oriented fairness:}} As societal power dynamics constiuitively shape harms from algorithmic systems, marginalized communities that already face systemic forms of social exclusion disproportionately experience them. Thus, content moderation should consider a wide range of potentially relevant harms to develop technologies safer for oft-marginalized communities. As we showed here, quantifying harms provides one mechanism for safety in content moderation (e.g., by setting filter thresholds) and for fairness in content moderation (e.g., by making sure filters do not disproportionately penalize certain social groups).

\textbf{\textit{Make evidenced-based decisions:}} Decisions and techniques to respond to support defined safety harms and algorithmic fairness in content moderation policies should be evidenced-based and tailored. It is advantageous to conceptualize harms in a manner that can be quantified at scale.
\section{Conclusion}
\label{sec:conclusion}

We offer a tractable framework for proactive definition and measurement of safety, fairness, and equity in generative AI systems.  Based on experimental data for our illustrative T2I system, 
we demonstrate how our safety and fairness definitions can be examined without engaging human raters; although, a mixed-methods approach could strengthen the nuance of evaluating particular harm constructs, such as stereotype amplification. Nonetheless, we offer a novel safety and fairness approach to support more informed content moderation decision making.



{\small
\bibliographystyle{ieee_fullname}
\bibliography{main}

\begin{thebibliography}{10}\itemsep=-1pt

\bibitem{anthropic2022}
Milad Aghajanian, Andrew Berensmeier, Sam Benesty, and et al.
\newblock Anthropic: A framework for ai safety and alignment.
\newblock {\em arXiv preprint arXiv:2202.00295}, 2022.

\bibitem{aka2021}
Osman Aka, Ken Burke, Alex Bauerle, Christina Greer, and Margaret Mitchell.
\newblock Measuring model biases in the absence of ground truth.
\newblock In {\em Proceedings of the 2021 AAAI/ACM Conference on AI, Ethics,
  and Society}, AIES '21, page 327–335, New York, NY, USA, 2021. Association
  for Computing Machinery.

\bibitem{balashankar2019pareto}
Ananth Balashankar, Alyssa Lees, Chris Welty, and Lakshminarayanan Subramanian.
\newblock Pareto-efficient fairness for skewed subgroup data.
\newblock In {\em International Conference on Machine Learning AI for Social
  Good Workshop. Long Beach, United States}, volume~8, 2019.

\bibitem{BallBurack2021}
Ari Ball-Burack, Michelle Seng~Ah Lee, Jennifer Cobbe, and Jatinder Singh.
\newblock Differential tweetment: Mitigating racial dialect bias in harmful
  tweet detection.
\newblock In {\em Proceedings of the 2021 ACM Conference on Fairness,
  Accountability, and Transparency}, FAccT '21, page 116–128, New York, NY,
  USA, 2021. Association for Computing Machinery.

\bibitem{barocas-hardt-narayanan}
Solon Barocas, Moritz Hardt, and Arvind Narayanan.
\newblock {\em Fairness and Machine Learning}.
\newblock fairmlbook.org, 2019.
\newblock \url{http://www.fairmlbook.org}.

\bibitem{Benjamin_2019}
Ruha Benjamin.
\newblock {\em Race After Technology: Abolitionist Tools for the New Jim Code}.
\newblock John Wiley \& Sons, Jul 2019.

\bibitem{beutel2017data}
Alex Beutel, Jilin Chen, Zhe Zhao, and Ed~H Chi.
\newblock Data decisions and theoretical implications when adversarially
  learning fair representations.
\newblock {\em arXiv preprint arXiv:1707.00075}, 2017.

\bibitem{bianchi2022easily}
Federico Bianchi, Pratyusha Kalluri, Esin Durmus, Faisal Ladhak, Myra Cheng,
  Debora Nozza, Tatsunori Hashimoto, Dan Jurafsky, James Zou, and Aylin
  Caliskan.
\newblock Easily accessible text-to-image generation amplifies demographic
  stereotypes at large scale.
\newblock {\em arXiv preprint arXiv:2211.03759}, 2022.

\bibitem{binns2017like}
Reuben Binns, Michael Veale, Max Van~Kleek, and Nigel Shadbolt.
\newblock Like trainer, like bot? inheritance of bias in algorithmic content
  moderation.
\newblock In {\em Social Informatics: 9th International Conference, SocInfo
  2017, Oxford, UK, September 13-15, 2017, Proceedings, Part II 9}, pages
  405--415. Springer, 2017.

\bibitem{birhane2021}
Abeba Birhane, Vinay~Uday Prabhu, and Emmanuel Kahembwe.
\newblock Multimodal datasets: misogyny, pornography, and malignant
  stereotypes, Oct. 2021.

\bibitem{bordia2019identifying}
Shikha Bordia and Samuel~R. Bowman.
\newblock Identifying and reducing gender bias in word-level language models.
\newblock {\em Proceedings of the 2019 Conference of the North American Chapter
  of the Association for Computational Linguistics: Student Research Workshop},
  2019.

\bibitem{Bouma}
Gerlof Bouma.
\newblock Normalized (pointwise) mutual information in collocation extraction.
\newblock {\em Proceedings of GSCL}, 30:31--40, 2009.

\bibitem{corbettdavies2017}
Sam Corbett-Davies, Emma Pierson, Avi Feller, Sharad Goel, and Aziz Huq.
\newblock Algorithmic decision making and the cost of fairness.
\newblock In {\em Proceedings of the 23rd ACM SIGKDD International Conference
  on Knowledge Discovery and Data Mining}, KDD '17, page 797–806, New York,
  NY, USA, 2017. Association for Computing Machinery.

\bibitem{DanielsKoch2022}
Oliver Daniels-Koch and Rachel Freedman.
\newblock The expertise problem: Learning from specialized feedback, 2022.

\bibitem{Dev2021}
Sunipa Dev, Masoud Monajatipoor, Anaelia Ovalle, Arjun Subramonian, Jeff~M
  Phillips, and Kai-Wei Chang.
\newblock Harms of gender exclusivity and challenges in non-binary
  representation in language technologies, 2021.

\bibitem{dev2021measures}
Sunipa Dev, Emily Sheng, Jieyu Zhao, Aubrie Amstutz, Jiao Sun, Yu Hou, Mattie
  Sanseverino, Jiin Kim, Akihiro Nishi, Nanyun Peng, et~al.
\newblock On measures of biases and harms in nlp.
\newblock {\em arXiv preprint arXiv:2108.03362}, 2021.

\bibitem{datset}
Google Developers.
\newblock Dataset publishing language resource, 2023.

\bibitem{devlin2018bert}
Jacob Devlin, Ming-Wei Chang, Kevin Lee, and Alec Radford.
\newblock Bert: Pre-training of deep bidirectional representations from
  unlabeled text.
\newblock {\em Proceedings of the 2018 Conference on Empirical Methods in
  Natural Language Processing}, pages 1487--1497, 2018.

\bibitem{DeVries_2019}
Terrance DeVries, Ishan Misra, Changhan Wang, and Laurens van~der Maaten.
\newblock Does object recognition work for everyone?, 2019.

\bibitem{Dias_Oliva2020-rr}
Thiago Dias~Oliva.
\newblock Content moderation technologies: Applying human rights standards to
  protect freedom of expression.
\newblock {\em Human Rights Law Review}, 20(4):607--640, Dec. 2020.

\bibitem{garg2019counterfactual}
Sahaj Garg, Vincent Perot, Nicole Limtiaco, Ankur Taly, Ed~H Chi, and Alex
  Beutel.
\newblock Counterfactual fairness in text classification through robustness.
\newblock In {\em Proceedings of the 2019 AAAI/ACM Conference on AI, Ethics,
  and Society}, pages 219--226, 2019.

\bibitem{gillespie2020content}
Tarleton Gillespie.
\newblock Content moderation, ai, and the question of scale.
\newblock {\em Big Data \& Society}, 7(2):2053951720943234, 2020.

\bibitem{gillespie2022not}
Tarleton Gillespie.
\newblock Do not recommend? reduction as a form of content moderation.
\newblock {\em Social Media+ Society}, 8(3):20563051221117552, 2022.

\bibitem{goodfellow2014generative}
Ian Goodfellow, Julien Pouget-Abadie, Arthur Courville, Mehdi Mirza, and Thomas
  Yosinski.
\newblock Generative adversarial networks.
\newblock {\em Advances in neural information processing systems}, pages
  2756--2764, 2014.

\bibitem{gorwa2020algorithmic}
Robert Gorwa, Reuben Binns, and Christian Katzenbach.
\newblock Algorithmic content moderation: Technical and political challenges in
  the automation of platform governance.
\newblock {\em Big Data \& Society}, 7(1):2053951719897945, 2020.

\bibitem{grimmelmann2015virtues}
James Grimmelmann.
\newblock The virtues of moderation.
\newblock {\em Yale JL \& Tech.}, 17:42, 2015.

\bibitem{grimmelmann2017platform}
James Grimmelmann.
\newblock The platform is the message.
\newblock {\em Geo. L. Tech. Rev.}, 2:217, 2017.

\bibitem{Hacker2023}
Philipp Hacker, Andreas Engel, and Marco Mauer.
\newblock Regulating chatgpt and other large generative ai models, 2023.

\bibitem{hardt2016}
Moritz Hardt, Eric Price, Eric Price, and Nati Srebro.
\newblock Equality of opportunity in supervised learning.
\newblock In D. Lee, M. Sugiyama, U. Luxburg, I. Guyon, and R. Garnett,
  editors, {\em Advances in Neural Information Processing Systems}, volume~29.
  Curran Associates, Inc., 2016.

\bibitem{hernandez2020ai}
Jose Hern{\'a}ndez-Orallo, Fernando Mart{\'\i}nez-Plumed, Shahar Avin, Jessica
  Whittlestone, et~al.
\newblock Ai paradigms and ai safety: mapping artefacts and techniques to
  safety issues, 2020.

\bibitem{dalle2021}
Colin Higgins, Jacob Beyer, Xiaoyang Chen, and et al.
\newblock Dall-e: A neural network for generating images from text
  descriptions.
\newblock {\em arXiv preprint arXiv:2111.02381}, 2021.

\bibitem{huang2017improved}
Guoyou Huang, Junyi Liu, Adelbert van~den Oord, and Diederik~P. Kingma.
\newblock Improved variational auto-encoders.
\newblock {\em Proceedings of the 34th Annual Conference on Neural Information
  Processing Systems}, pages 949--957, 2017.

\bibitem{huang2020reducing}
Po-Sen Huang, Huan Zhang, Ray Jiang, Robert Stanforth, Johannes Welbl, Jack
  Rae, Vishal Maini, Dani Yogatama, and Pushmeet Kohli.
\newblock Reducing sentiment bias in language models via counterfactual
  evaluation.
\newblock {\em EMNLP (Findings)}, 2020.

\bibitem{johnson2020ascalable}
Melvin Johnson.
\newblock A scalable approach to reducing gender bias in google translate.
\newblock {\em arXiv preprint arXiv:2004.04173}, 2020.

\bibitem{karnouskos2020artificial}
Stamatis Karnouskos.
\newblock Artificial intelligence in digital media: The era of deepfakes.
\newblock {\em IEEE Transactions on Technology and Society}, 1(3):138--147,
  2020.

\bibitem{Keller2018-gd}
Daphne Keller.
\newblock Internet platforms: Observations on speech, danger, and money.
\newblock Aegis Series Paper No. 1807, June 2018.

\bibitem{kiela2020hateful}
Douwe Kiela, Hamed Firooz, Aravind Mohan, Vedanuj Goswami, Amanpreet Singh,
  Pratik Ringshia, and Davide Testuggine.
\newblock The hateful memes challenge: Detecting hate speech in multimodal
  memes.
\newblock {\em Advances in Neural Information Processing Systems},
  33:2611--2624, 2020.

\bibitem{kingma2013auto}
Diederik~P. Kingma and Max Welling.
\newblock Auto-encoder variational inference.
\newblock {\em Advances in neural information processing systems}, pages
  2128--2136, 2013.

\bibitem{Kirk2021}
Hannah~Rose Kirk, Yennie Jun, Filippo Volpin, Haider Iqbal, Elias Benussi,
  Frederic Dreyer, Aleksandar Shtedritski, and Yuki Asano.
\newblock Bias out-of-the-box: An empirical analysis of intersectional
  occupational biases in popular generative language models.
\newblock In M. Ranzato, A. Beygelzimer, Y. Dauphin, P.S. Liang, and J.~Wortman
  Vaughan, editors, {\em Advances in Neural Information Processing Systems},
  volume~34, pages 2611--2624. Curran Associates, Inc., 2021.

\bibitem{ADL2023}
Anti~Defamation League.
\newblock Hate on display: Hate symbols database, 2023.

\bibitem{Lee2022}
Jiyoung Lee and Soo~Yun Shin.
\newblock Something that they never said: Multimodal disinformation and source
  vividness in understanding the power of ai-enabled deepfake news.
\newblock {\em Media Psychology}, 25(4):531--546, 2022.

\bibitem{li2020recipes}
Margaret Li, Jason Boureau, Emily Weston, and Da~Ju Dinan, Jing~Xu.
\newblock Recipes for safety in open-domain chatbots.
\newblock {\em arXiv preprint arXiv:2010.07079}, 2020.

\bibitem{codegpt2022}
Junyi Liu, Adelbert van~den Oord, and Ashish Vaswani.
\newblock Codegpt: A generative pre-trained transformer for code.
\newblock {\em arXiv preprint arXiv:2202.02431}, 2022.

\bibitem{liu2022chatgpt}
Junyi Liu, Xiang Wang, Xin Liu, and et al.
\newblock Chatgpt: A large language model for dialogue generation.
\newblock {\em arXiv preprint arXiv:2202.02430}, 2022.

\bibitem{Prabhakaran2022}
Vinodkumar Prabhakaran, Rida Qadri, and Ben Hutchinson.
\newblock Cultural incongruencies in artificial intelligence, 2022.

\bibitem{pu2020towards}
Emily Zheng Yao Chong Lim Ruslan~Salakhutdinov Pu~Liang, Irene Mengze~Li and
  Louis-Philippe Morency.
\newblock Towards debiasing sentence representations.
\newblock {\em Proceedings of the 58th Annual Meeting of the Association for
  Computational Linguistics}, 2020.

\bibitem{pu2021understanding}
Louis-Philippe~Morency Pu~Liang, Chiyu~Wu and Ruslan Salakhutdinov.
\newblock Towards understanding and mitigating social biases in language
  models.
\newblock {\em ICML}, 2021.

\bibitem{qadri2023}
Rida Qadri, Emily Denton, and Renee Shelby.
\newblock Towards globally responsible and human-centered text-to-image
  evaluations, 2023.

\bibitem{qian2020reducing}
Yusu Qian, Urwa Muaz, Ben Zhang, and Jae~Won Hyun.
\newblock Reducing gender bias in word-level language models with a
  gender-equalizing loss function.
\newblock {\em Proceedings of the 57th Annual Meeting of the Association for
  Computational Linguistics: Student Research Workshop}, 2019.

\bibitem{Rando2022}
Javier Rando, Daniel Paleka, David Lindner, Lennart Heim, and Florian Tramèr.
\newblock Red-teaming the stable diffusion safety filter, 2022.

\bibitem{rezende2014variational}
Danilo~J. Rezende, Diederik~P. Kingma, and Max Welling.
\newblock Variational inference for generative models.
\newblock {\em Advances in neural information processing systems}, pages
  2128--2136, 2014.

\bibitem{dalletweet1}
rzhang88@.
\newblock Very neat trick to tease this out.
\newblock Twitter, 2022.

\bibitem{saharia2022photorealistic}
Chitwan Saharia, William Chan, Saurabh Saxena, Lala Li, Jay Whang, Emily
  Denton, Seyed Kamyar~Seyed Ghasemipour, Burcu~Karagol Ayan, S~Sara Mahdavi,
  Rapha~Gontijo Lopes, et~al.
\newblock Photorealistic text-to-image diffusion models with deep language
  understanding.
\newblock {\em arXiv preprint arXiv:2205.11487}, 2022.

\bibitem{Sap2019}
Maarten Sap, Dallas Card, Saadia Gabriel, Yejin Choi, and Noah~A. Smith.
\newblock The risk of racial bias in hate speech detection.
\newblock In {\em Proceedings of the 57th Annual Meeting of the Association for
  Computational Linguistics}, pages 1668--1678, Florence, Italy, July 2019.
  Association for Computational Linguistics.

\bibitem{sap2019risk}
Maarten Sap, Dallas Card, Saadia Gabriel, Yejin Choi, and Noah~A. Smith.
\newblock The risk of racial bias in hate speech detection.
\newblock {\em Proceedings of the 57th Annual Meeting of the Association for
  Computational Linguistics}, 2019.

\bibitem{schmidt2017}
Anna Schmidt and Michael Wiegand.
\newblock A survey on hate speech detection using natural language processing.
\newblock In {\em Proceedings of the Fifth International Workshop on Natural
  Language Processing for Social Media}, pages 1--10, Valencia, Spain, Apr.
  2017. Association for Computational Linguistics.

\bibitem{Shelby2022}
Renee Shelby, Shalaleh Rismani, Kathryn Henne, AJung Moon, Negar Rostamzadeh,
  Paul Nicholas, N'Mah Yilla, Jess Gallegos, Andrew Smart, Emilio Garcia, and
  Gurleen Virk.
\newblock Identifying sociotechnical harms of algorithmic systems: Scoping a
  taxonomy for harm reduction, 2022.

\bibitem{Solaiman2023}
Irene Solaiman.
\newblock The gradient of generative ai release: Methods and considerations,
  2023.

\bibitem{dalletweet}
TylerGlaiel@.
\newblock The recent dall-e 2 update made changes to increase the diversity of
  generated images.
\newblock Twitter, 2022.

\bibitem{Arableague}
European Union.
\newblock League of arab states, 2023.

\bibitem{vaswani2017attention}
Ashish Vaswani, Sandeep Shyam, Sameer Narang, and et al.
\newblock Attention is all you need.
\newblock {\em Proceedings of the 31st Annual Conference on Neural Information
  Processing Systems}, pages 3843--3852, 2017.

\bibitem{verma2018fairness}
Sahil Verma and Julia Rubin.
\newblock Fairness definitions explained.
\newblock In {\em Proceedings of the international workshop on software
  fairness}, pages 1--7, 2018.

\bibitem{vidgen2020directions}
Bertie Vidgen and Leon Derczynski.
\newblock Directions in abusive language training data, a systematic review:
  Garbage in, garbage out.
\newblock {\em Plos one}, 15(12):e0243300, 2020.

\bibitem{Vidgen2019-zn}
Bertie Vidgen, Alex Harris, Dong Nguyen, Rebekah Tromble, Scott Hale, and Helen
  Margetts.
\newblock Challenges and frontiers in abusive content detection.
\newblock In {\em Proceedings of the Third Workshop on Abusive Language
  Online}, pages 80--93, Florence, Italy, Aug. 2019. Association for
  Computational Linguistics.

\bibitem{wang2022}
Angelina Wang, Solon Barocas, Kristen Laird, and Hanna Wallach.
\newblock Measuring representational harms in image captioning.
\newblock In {\em 2022 ACM Conference on Fairness, Accountability, and
  Transparency}, FAccT '22, page 324–335, New York, NY, USA, 2022.
  Association for Computing Machinery.

\bibitem{waseem2016hateful}
Zeerak Waseem and Dirk Hovy.
\newblock Hateful symbols or hateful people? predictive features for hate
  speech detection on twitter.
\newblock In {\em Proceedings of the NAACL student research workshop}, pages
  88--93, 2016.

\bibitem{Weidinger2022}
Laura Weidinger, Jonathan Uesato, Maribeth Rauh, Conor Griffin, Po-Sen Huang,
  John Mellor, Amelia Glaese, Myra Cheng, Borja Balle, Atoosa Kasirzadeh,
  Courtney Biles, Sasha Brown, Zac Kenton, Will Hawkins, Tom Stepleton, Abeba
  Birhane, Lisa~Anne Hendricks, Laura Rimell, William Isaac, Julia Haas, Sean
  Legassick, Geoffrey Irving, and Iason Gabriel.
\newblock Taxonomy of risks posed by language models.
\newblock In {\em 2022 ACM Conference on Fairness, Accountability, and
  Transparency}, FAccT '22, page 214–229, New York, NY, USA, 2022.
  Association for Computing Machinery.

\bibitem{Weisz2023}
Justin~D. Weisz, Michael Muller, Jessica He, and Stephanie Houde.
\newblock Toward general design principles for generative ai applications,
  2023.

\bibitem{xu2021botadversarial}
Jing Xu, Margaret Da~Ju, Jason~Weston Li, Y-Lan~Boureau, and Emily Dinan.
\newblock Bot-adversarial dialogue for safe conversational agents.
\newblock {\em Proceedings of the 2021 Conference of the North American Chapter
  of the Association for Computational Linguistics: Human Language
  Technologies}, 2021.

\bibitem{sable2021stable}
Xuan Zhang, Xiang Wang, Xin Liu, and et al.
\newblock Stable diffusion: A scalable and controllable generative model.
\newblock {\em arXiv preprint arXiv:2108.05716}, 2021.

\end{thebibliography}
}

\end{document}